\documentclass[10pt,twocolumn,letterpaper]{article}

\usepackage{cvpr}
\usepackage{times}
\usepackage{epsfig}
\usepackage{graphicx}
\usepackage{amsmath}
\usepackage{amssymb}

\usepackage{algorithm, algorithmic}
\usepackage{multirow}
\usepackage{booktabs}   
\usepackage{amssymb} 
\usepackage{bm}
\usepackage{authblk}

\usepackage[pagebackref=true,breaklinks=true,letterpaper=true,colorlinks,bookmarks=false]{hyperref}

\cvprfinalcopy 

\def\cvprPaperID{****} 

\ifcvprfinal\fi

\begin{document}

\title{Self-Learning with Rectification Strategy for Human Parsing}

\title{Self-Learning with Rectification Strategy for Human Parsing}


\author{Tao Li$^1$,
Zhiyuan Liang$^1$,
Sanyuan Zhao\thanks{Corresponding author: \textit{Sanyuan Zhao}.
This work was supported in part by the National Natural Science Foundation of China (61902027),
the Beijing Natural Science Foundation (4182056),
the CCF-Tencent Open Fund, Zhijiang Lab's International Talent Fund for Young Professionals.
}~~$^{1}$, Jiahao Gong$^1$, Jianbing Shen$^{2,1}$\\
$^1$\small Beijing Lab of Intelligent Information Technology, School of Computer Science, Beijing Institute of Technology, China\\
$^2$\small Inception Institute of Artificial Intelligence, Abu Dhabi, UAE\\
{\tt\small \{tao.li, liangzhiyuan, zhaosanyuan, gongjiahao\}@bit.edu.cn, shenjianbingcg@gmail.com \\
}
}

\maketitle

\begin{abstract}
In this paper, we solve the sample shortage problem in the human parsing task. We begin with the self-learning strategy, which generates pseudo-labels for unlabeled data to retrain the model. However, directly using noisy pseudo-labels will cause error amplification and accumulation. Considering the topology structure of human body, we propose a trainable graph reasoning method that establishes internal structural connections between graph nodes to correct two typical errors in the pseudo-labels, i.e., the global structural error and the local consistency error. For the global error, we first transform category-wise features into a high-level graph model with coarse-grained structural information, and then decouple the high-level graph to reconstruct the category features. The reconstructed features have a stronger ability to represent the topology structure of the human body. Enlarging the receptive field of features can effectively reducing the local error. We first project feature pixels into a local graph model to capture pixel-wise relations in a hierarchical graph manner, then reverse the relation information back to the pixels. With the global structural and local consistency modules, these errors are rectified and confident pseudo-labels are generated for retraining. Extensive experiments on the LIP and the ATR datasets demonstrate the effectiveness of our global and local rectification modules. Our method outperforms other state-of-the-art methods in supervised human parsing tasks.
\end{abstract}


\vspace{-2mm}
\section{Introduction}

\begin{figure*}
    \centering
    \includegraphics[width = \textwidth]{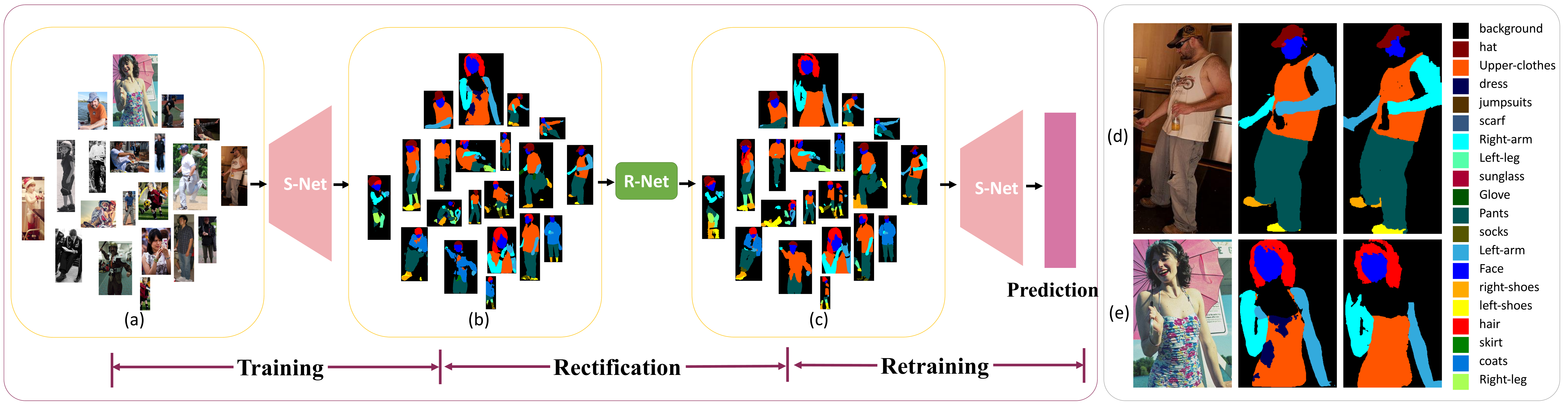}
    \caption{The left column shows our self-learning process.
    The sub-figures (a), (b), (c) respectively denote the partially labeled images, the predicted masks, and the rectified high-quality pseudo-labels.
    The right columns show the common errors in human parsing, including global structure errors (d) and local consistency errors (e).
    We propose the retraining strategy with our rectification framework to correct the predicted errors in segmentation network, and retrain the network with labeled data and the generated high-quality pseudo-labeled samples.
    }
    \label{fig:1}
    \vspace{-0.4cm}
\end{figure*}

Human parsing, a sub-task of semantic segmentation, aims to understand human-body parts on the pixel level.
In particular, human parsing is characterized by utilizing the structure of the human body.
It has been widely applied in human-computer interaction \cite{lin2016virtual}, human behavior understanding \cite{wang2013approach,fan2019understanding,Zhou2020human}, security monitoring \cite{zhou2016sparseness,foroughi2008intelligent} etc.
For deep neural network-based algorithms, they learn complex information from a large amount of labeled samples to boost performance.
But collecting accurate and fine-grained labeling for human parsing is very expensive and needs massive human labor.
With insufficient training samples, weakly-supervised and semi-supervised methods are proposed to address this issue.
Most existing algorithms adopt the human posture or skeleton key points as a supplement to human parsing~\cite{fang2018weakly,lin2019cross}.
However, these algorithms require extra computing resources of human pose or key points, which are usually unavailable in real cases and may introduce new errors~\cite{zhang2019pose2seg}.



To enlarge the number of samples for training, we regard the predicted masks of unlabeled images as their pseudo-labels and retrain the segmentation network.
But these noisy pseudo-labels contain many errors.
If the network is blindly confident of its incorrect predictions, the error will be amplified during retraining~\cite{reichart2007self}.
Thus, the technical bottleneck is how to autonomously correct the errors of pseudo-labels.
For severely noisy pseudo-labels in self-training, some label denoising algorithms are proposed. 
The transition matrix is adopted to capture the transition probability between the noisy label and the true label~\cite{hendrycks2018using,patrini2017making}, and the extra linear network is added to evaluate the noise \cite{goldberger2016training,sukhbaatar2014training}.
However, these algorithms conducting experiments on the simple MNIST dataset are hard to learn the complex noise in human parsing.
We propose a new rectification network to detect and correct the errors of pseudo-labels by graph reasoning.

There are two main types of predicted errors in human parsing \cite{nie2018mutual,luo2018macro}, \emph{i.e.}, 
the global structure error \cite{gong2017look} and the local consistency error (Fig.~\ref{fig:1}).
The first type is the inter-part error in the human body.
For example, in Fig. \ref{fig:1}~(d), the left-arm is incorrectly predicted as the right-arm.
It is mainly caused by reasoning errors on the human structural level.
The second type is the intra-part error such as the spotted noise in an area of a certain category.
As shown in Fig. \ref{fig:1}~(e), some of the pixels belonging to the upper-clothes are predicted as the dress due to the limited receptive field of the networks.
Luo \emph{et al.} \cite{luo2018macro} proposed macro and micro discriminators respectively for the two types of errors, but the adversarial strategy leads to unstable training process with interminable training time.
Nie \emph{et al.}~\cite{nie2018mutual} proposed to jointly learn human parsing and pose estimation, but their algorithm needs doubled computing resources and multiplied training time.
Because of the topology structure of human body, it is natural to build a graph model for it and perform graph reasoning to solve the structural errors.

Based on the idea of pseudo-label denoising in self-learning, we propose a new Graph Rectification Network that learns the hierarchical structure of human body and solves the above-mentioned errors by constructing a dual graph reasoning framework.
To deal with the global structural error, we introduce a Global Structure Module for graph reasoning.
We first transform the grid features into a low-level semantic graph, where each node explicitly represents a certain body-part category (\emph{e.g.}, left leg, right arm, and coat).
Then the low-level graph is aggregated into a high-level graph to implicitly represent coarse-grained human parts (\emph{e.g.}, upper limbs, lower limbs and clothes).
The hierarchical knowledge is transferred back to each category by decoupling the high-level graph to reconstruct a new low-level graph.
The reconstructed low-level graph are more discriminative on confusing problems caused by similar appearances.
As for the local consistency error, we build a Local Consistency Module to process spatial pixels.
It indirectly defines the relationship between pixels in an enlarged receptive field.
Meanwhile, the globally structural information works as the auxiliary knowledge to solve the local errors.
By guiding the global structural relationship and enlarging the local receptive field, the local consistency error is corrected.
Thus, we combine the global graph with the local graph to boost performance of the fine-grained segmentation.
The dual hierarchical structural information acts as a structural attention to rectify the pseudo-labels.

Our paper has the following main contributions.
We first propose a semi-supervised training framework named as Graph Rectification Network to solve the sample insufficiency problem in human parsing.
Then, a Global Structure Module and a Local Consistency Module are designed to rectify the global and local errors of pseudo-labels by constructing the graph reasoning models on different semantic levels.
Finally, our method can be applied to different baseline algorithms with state-of-the-art performance.

\section{Related Work}

{\bf Human Parsing.}
Human parsing is a subtask of semantic segmentation, but with the particular structure constraint on the human body \cite{wang2019learning,Zhou2020motionsegment}.
Gong \emph{et al.} \cite{gong2017look} introduced the pseudo pose loss as an auxiliary constraint to assist human parsing.
The works of \cite{zhang2019pose2seg,xia2017joint,nie2018mutual} jointly trained pose estimation and human parsing networks to improve the performance of both tasks.
Wang \emph{et al.} \cite{wang2019learning} adopted the bottom-up and top-down hierarchical human body structure to reason human part segmentation and achieved the state-of-the-art performance.
In our work, we exploit the hierarchical spirit in our graph reasoning modules.
Through trainable aggregation strategy, we transform the low-level nodes corresponding to different human parts to implicit high-level nodes.
Then we revert them to low-level nodes through a trainable decoupling strategy, which makes each low-level node carry the structural knowledge.

\begin{figure*}
    \centering
    \includegraphics[width = 1\textwidth]{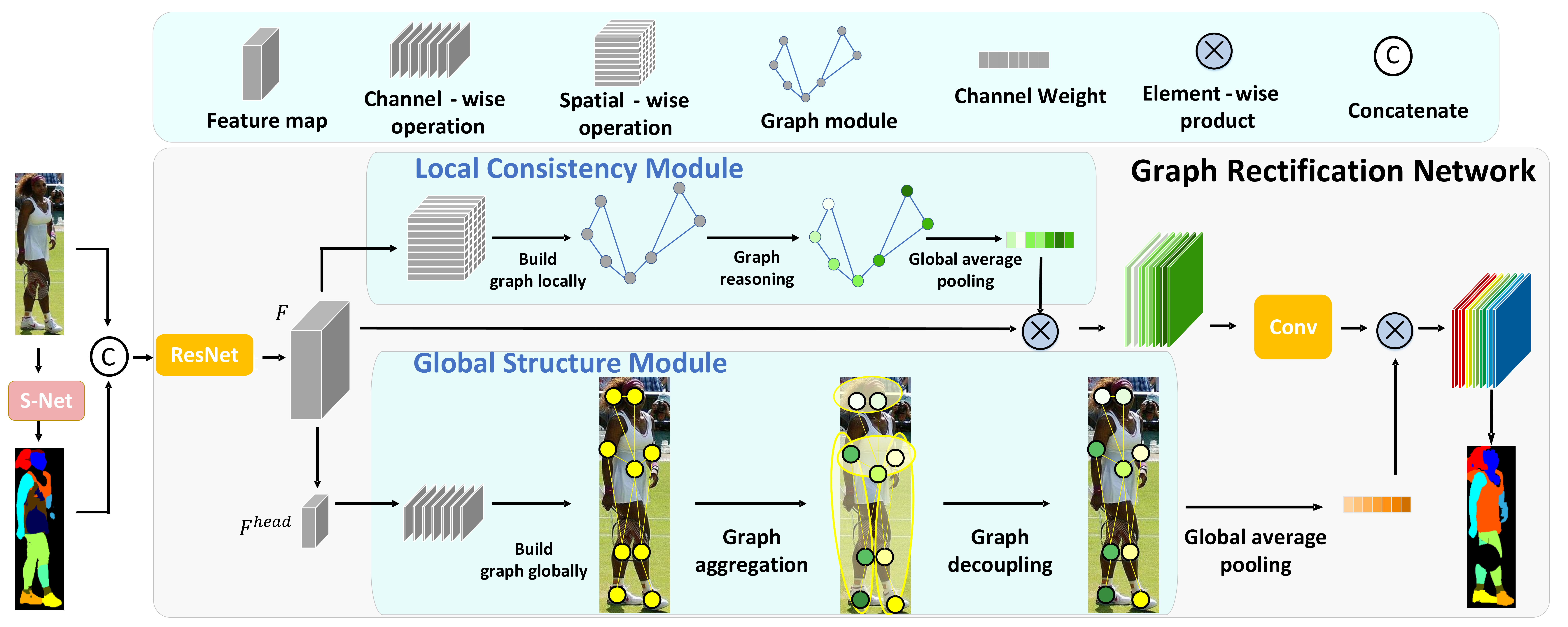}
    \caption{Illustration of our rectification network.
    We concatenate image and predicted mask as input for the rectification network. Then we perform the global structure and local consistency rectification for the two types of errors. The S-Net denotes the segmentation network.}
    \label{fig:2}
    \vspace{-0.4cm}
\end{figure*}

{\bf Semi-supervised learning.}
In image and video segmentation, sufficient and accurate annotations are helpful for network training \cite{liang2015deep,liang2018look,chen2014detect,Hu2020gcngp,DBLP:journals/pami/WangSPY19,DBLP:journals/pami/WangSYP18}.
However, the annotation work is time-consuming and requires amounts of human resources. Existing datasets may not satisfy the demand of network training.
Recently, semi-supervised and weakly-supervised methods have emerged.
To fuse the specific structural information of the human body, \cite{fang2018weakly,lin2019cross} adopted pose predictions of the target domain to assist the segmentation of the source domain.
\cite{gong2019graphonomy} proposed to augment the number of samples from other datasets by matching the corresponding relationship among categories via the graph model.
Generating pseudo-labels for wild data is an efficient strategy for data augmentation \cite{lee2013pseudo,ding2018semi,tanaka2018joint}.
However, the self-learning strategy with noisy pseudo-labels may result in error amplification and accumulation.
To address it, many label denoising algorithms have been proposed~\cite{han2019deep}.
The transition matrix was adopted to capture the transition probability between the noisy label and the true label~\cite{hendrycks2018using,patrini2017making,goldberger2016training,sukhbaatar2014training}.
However, it is hard to learn the transition matrix due to the agnosticism of neural networks.

We follow the idea of pseudo-label denoising in self-learning and propose the Graph Rectification Network, which learns the hierarchical structure of the human body and assigns the global and local information to each graph node.
This structural information helps to rectify the global structural error and the local consistency error happened when the training samples are not sufficient.
Thus more confident pseudo-labels are generated for expanding the training set by semi-supervised learning.


{\bf Graph reasoning.}
Rex \emph{et al.} \cite{ying2018hierarchical} proposed a differentiable pooling for graph representation by aggregating multiple nodes with similar relations into one node to obtain an advanced graph model.
Gao \emph{et al.} \cite{gao2019graph} further proposed a graph U-Net algorithm, which allows the graph information to be aggregated and spread out, resulting in a graph model with stronger representation ability.
Because of the effectiveness of graph reasoning, it has been quickly applied in various fields, such as human re-id \cite{ristani2018features,Ye2020reid,Ye2020embed}, motion recognition \cite{si2018skeleton,wang2018videos,DBLP:conf/iccv/WangLSC019}, human-object relation reasoning \cite{wang2019learning,DBLP:conf/eccv/QiWJSZ18,DBLP:conf/iccv/ShenWLSLX019,Wang2020parsing}, and multi-label classification problem \cite{chen2019multi}.
In visual semantic segmentation, \cite{chen2019graph,li2018beyond} transformed the grid Euclidean data into the structural graph model, and performed graph reasoning on it.
The human body is a typical topology structure with skeleton key point constraints (\emph{e.g.}, the left arm and left foot should be on one side of the body).
So graph reasoning is natural to be adopted in this task.
Gong \emph{et al.} \cite{gong2019graphonomy} took full advantage of the labeled datasets by modeling the relations between different domains with a pre-defined adjacency matrix.
Our work constructs a global graph model by mapping the category-wise features into graph nodes and aggregates these nodes into a higher-level graph to acquire hierarchical structure knowledge of the human body.
\section{Method}
\label{section_3}
\subsection{Overview}

Our pipeline is shown in Fig. \ref{fig:2}. We put images into the Segmentation Network (S-Net) to get the predicted masks.
Then we concatenate the images and predicted masks and deliver them into our rectification network to get the globally and locally rectified masks.
We use the rectified masks to retrain the S-Net.
The Rectification Network (R-Net) consists of three parts, that are the backbone, the Global Structure Module (GSM) and the Local Consistency Module (LCM).
The GSM module is introduced in \ref{section_3.2} and \ref{section_3.3} including graph construction and hierarchical graph reasoning.
The LCM module is introduced in \ref{section_3.4}.
We introduce the integration method of the GSM and LCM in \ref{section_3.5}, and our semi-supervised training strategy in \ref{section_3.6}.

The global error is the predicted error of the entire human parts, such as confusing left and right arms.
The local error means that pixels belonging to one human part may be assigned to two or more categories (\emph{e.g.} the pixels belonging to dress are assigned to dress and upper-clothes).
The GSM and LCM are different in the way the graph is built for different optimization objectives.

\subsection{Build Global Graph Model Explicitly}
\label{section_3.2}
To parse the hard samples such as complicated postures or limbs absence, we propose a global graph model to explicitly represent the human-body features, and then optimize the model through an implicit graph model hierarchical reasoning.
As shown in Fig. \ref{fig:2}, the S-Net generates features of noisy masks $ {\bf{F}}^{head} \in \mathbb R^{c\times w \times h}$, where $c$, $w$ and $h$  represent number of categories, the width and the height of the feature map, respectively.

To semantically model the features, we perform the one-to-one correspondence between the channels of the feature map and the nodes of the graph model.
We globally transform the feature ${\bf F}^{head}$ into a semantically low-level undirected graph model ${\bf U}_g^{(low)}=\{({\bf X}_g^{(low)}, { \bf A}_g^{(low)})\}$, where ${ \bf X}_g^{(low)} \in \mathbb R^{c\times d}$ denotes the eigenvectors of graph nodes, $d$ denotes the dimension of each eigenvector, ${\bf A}_g^{(low)} \in \mathbb R^{c \times c}$ denotes the adjacency matrix of the graph model and $c$ equals to the category number of the dataset.
${\bf A}_{g(i,j)}^{(low)}$ denotes the spatial adjacency between the class $i$ and the class $j$.
For the $i$-th channel of ${\bf F}^{head}$ corresponding to the $i$-th category, \emph{i.e.}, ${\bf F}_i^{head}\in \mathbb R^{w\times h}$, we transform it to a graph node as Eq. (\ref{eqn:1}):
\begin{equation}
   \setlength{\abovedisplayskip}{3pt}
   \setlength{\belowdisplayskip}{3pt}
    {{\bf X}_g}_i^{(low)}=\varphi_g({\bf F}_i^{head})\times {\bf \Omega}_i \in \mathbb R^{1 \times d},
    \label{eqn:1}
\end{equation}
where $\varphi_g ({\bf F})$ means reshaping the matrix ${\bf F}$ from $\mathbb R^{w\times h}$ to $\mathbb R^{1 \times (wh)}$, ${\bf \Omega}_i \in \mathbb R^{(wh) \times d}$ is the projection matrix of the $i$-th category, ${{\bf X}_g}_i^{(low)}$ is the eigenvector of the $i$-th node, and $\times$ denotes the matrix multiplication.
${\bf \Omega}_i$ is a learnable parameter and can be trained end-to-end.
The low-level graph model ${\bf U}_g^{(low)}$ contains the representation of human-body parts and the relations among them.

\subsection{Graph Model Aggregation/Decoupling}
\label{section_3.3}
Because the human body structure is hierarchical \cite{wang2019learning}, \emph{e.g.}, the upper body includes the head, the upper torso and the arms, the lower body includes the legs and the feet, the clothes include the coat, the pants and the dress, \emph{etc}.
We aggregate the low-level information of the relational graph nodes to perform higher-level structural information reasoning. In detail, we build the GSM to depict human structure information hierarchically and avoid the semantic structural errors by graph reasoning \cite{gong2019graphonomy}.


For the low-level graph model ${\bf U}_g^{(low)}$, we use Eq. (\ref{eqn:2}) to perform graph reasoning for information propagation of nodes.
\begin{equation}
   \setlength{\abovedisplayskip}{3pt}
   \setlength{\belowdisplayskip}{3pt}
    {\bf Z}_g^{(low)} = {\bf A}_g^{(low)} {\bf X}_g^{(low)} {\bf W}^{(low)} \in \mathbb R^{n_{low}\times d},
    \label{eqn:2}
\end{equation}
where ${\bf Z}_g^{(low)}$ is the node features after low-level graph reasoning, ${ \bf W}^{(low)}$ is the trainable weight matrix and $n_{low}$ denotes the number of nodes in the low-level graph model.

We propose a trainable aggregation matrix ${\bf C}^{(agg)}$ to convert the low-level graph model to the high-level graph model.
Then we obtain the high-level graph model ${\bf U}^{(high)}=\{({\bf X}_g^{(high)},{\bf A}_g^{(high)})\}$ by the Eq. (\ref{eqn:3}) and (\ref{eqn:4}),
\begin{equation}
   \setlength{\abovedisplayskip}{3pt}
   \setlength{\belowdisplayskip}{3pt}
    {\bf X}_g^{(high)} = {{\bf C}^{(agg)}}^T {\bf X}_g^{(low)} \in \mathbb R^{n_{high}\times d},
    \label{eqn:3}
\end{equation}
\begin{equation}
   \setlength{\abovedisplayskip}{3pt}
   \setlength{\belowdisplayskip}{3pt}
    {\bf A}_g^{(high)} = {{\bf C}^{(agg)}}^T {\bf A}_g^{(low)} {\bf C}^{(agg)} \in \mathbb R^{n_{high}\times n_{high}},
    \label{eqn:4}
\end{equation}
where ${\bf X}^{(high)}$ and $n_{high}$ denote the feature of nodes and the number of nodes in the high-level graph model, respectively.
For the high-level graph model, similar to Eq. (\ref{eqn:2}), we perform message passing between nodes with Eq. (\ref{eqn:5}) to get a new graph representation with more discriminative power,
\begin{equation}
   \setlength{\abovedisplayskip}{3pt}
   \setlength{\belowdisplayskip}{3pt}
    {\bf Z}_g^{(high)} = {\bf A}_g^{(high)} {\bf X}_g^{(high)} {\bf W}^{(high)} \in \mathbb R^{n_{high}\times d}.
    \label{eqn:5}
\end{equation}

Specifically, we compute the aggregation matrix with Eq. (\ref{eqn:6}),
\begin{equation}
   \setlength{\abovedisplayskip}{3pt}
   \setlength{\belowdisplayskip}{3pt}
    {\bf C}_g^{(agg)} = {\bf A}_g^{(low)} {\bf X}_g^{(low)} {\bf V}^{(low)} \in \mathbb R^{n_{low} \times n_{high}},
    \label{eqn:6}
\end{equation}
where ${\bf V}^{(low)} \in \mathbb R^{d\times n_{high}}$ is the trainable weight matrix, and $n_{high}$ is pre-set to control the number of nodes in the high-level graph model.
The $n_{low}$ is equal to the number of categories $c$ here.
This strategy makes our aggregation matrix ${\bf C}_g^{(agg)}$ trainable and improves the aggregating ability from low-level categories to high-level categories.

After aggregating the low-level categories (\emph{e.g.}, head, arm, pants and so on) into the implicit high-level categories (\emph{e.g.}, torso and clothes), we next revert the aggregated high-level model back to the low-level model.
However, we cannot separate them by simply converting according to corresponding relations because they are one-to-many correspondence.
Thus we perform decoupling processing to revert the aggregated information to the body-part level.
After decoupling, a set of low-level graph nodes with stronger discriminative features is generated.
Similar to the process of aggregation, we compute the trainable decoupling matrix by Eq. (\ref{eqn:7}), and decouple the high-level graph model by Eq. (\ref{eqn:8}) and (\ref{eqn:9}) to generate the reverted eigenvectors of the low-level graph $\hat{\bf Z}_g^{(low)}$ and the adjacency matrix $\hat{\bf A}_g^{(low)}$:

\begin{equation}
   \setlength{\abovedisplayskip}{3pt}
   \setlength{\belowdisplayskip}{3pt}
    {\bf C}_g^{(dec)} = {\bf A}_g^{(high)} {\bf X}_g^{(high)} {\bf V}^{(high)} \in \mathbb R^{n_{high} \times n_{low}},
    \label{eqn:7}
\end{equation}
\begin{equation}
   \setlength{\abovedisplayskip}{3pt}
   \setlength{\belowdisplayskip}{3pt}
    \hat{\bf Z}_g^{(low)} = {{\bf C}^{(dec)}}^T {\bf Z}_g^{(high)} \in \mathbb R^{n_{low}\times d},
    \label{eqn:8}
\end{equation}
\begin{equation}
   \setlength{\abovedisplayskip}{3pt}
   \setlength{\belowdisplayskip}{3pt}
    \hat{\bf A}_g^{(low)} = {{\bf C}^{(dec)}}^T {\bf A}_g^{(high)} {\bf C}^{(dec)} \in \mathbb R^{n_{low}\times n_{low}}.
    \label{eqn:9}
\end{equation}

In this way, we obtain the reverted low-level graph model $\hat{\bf U}^{(low)}=\{(\hat{\bf Z}_g^{(low)},\hat{\bf A}_g^{(low)})\}$ containing the high-level body structure information.

We apply a skip connection to integrate the original low-level graph with the reverted low-level graph,
\begin{equation}
   \setlength{\abovedisplayskip}{3pt}
   \setlength{\belowdisplayskip}{3pt}
   {\bf Z}_g = \hat{\bf Z}_g^{(low)} + {\bf Z}_g^{(low)}
\end{equation}

We enhance the discrimination of the features by the trainable aggregating and decoupling strategy to correct the common global structural errors of human parsing, \emph{e.g.}, the confusion caused by the similar appearance of human parts.
The eigenvector ${\bf Z}_g\in \mathbb R^{c\times d}$ of the reasoned graph model contains strong representation ability.

To apply hierarchical graph model to original feature maps, we convert the graph nodes ${\bf Z}_g$ to channel weights of ${\bf F}^{head}$ through Eq. (\ref{eqn:11}):
\vspace{-0.1cm}
\begin{equation}
   \setlength{\abovedisplayskip}{3pt}
   \setlength{\belowdisplayskip}{3pt}
   \renewcommand\arraystretch{1.4}
\left\{
   \begin{array}{l}
   {\bm{\theta}}_g= softmax(GAP({\bf Z}_g)) \\
   {\bf F}_{rectified}^{head} = {\bm \theta}_g \odot {\bf F}^{head}
   \end{array}
\right.
   \label{eqn:11}
\end{equation}
where $GAP$ denotes the global average pooling operation and ${\bf a} \odot {\bf B}$ denotes the element-wise product between a vector ${\bf a}$ and a tensor $\mathbf{B}$, where
\begin{equation}
   \setlength{\abovedisplayskip}{3pt}
   \setlength{\belowdisplayskip}{3pt}
   ({\bf a} \odot {\bf B})_i = {\bf a}_i \times {\bf B}_i
\label{eqn:12}
\end{equation}

This strategy can increase the weights of categories with strong correlations and reduce the weights of the irrelevant categories.

\subsection{Local Consistency Module}
\label{section_3.4}
In human parsing, another type of semantic error is the local consistency error.
It means that pixels belonging to one human part may be assigned to two or more categories.
This problem can be solved by computing the relation between pixels in an enlarged receptive field~\cite{chen2014semantic,wang2018non}.
The non-local algorithm \cite{wang2018non} was proposed to calculate the spatial-wise correlation of all the pixels to obtain global semantic information, but the algorithm brings a large amount of computational redundancy.
Similarly, it is not practical to set each pixel as a graph node and perform the graph reasoning directly, because the computational cost is too high to calculate the nodes relations.


Thus we project pixel features to graph nodes in our LCM.
After projection, we indirectly obtain the relations between pixels by calculating the relations between nodes.
By avoiding the local error, the performance can be improved through the enlarged receptive field within the controllable amount of calculation.

We perform projection from the input feature map ${\bf F}\in \mathbb R^{c'\times w' \times h'}$ to a local graph model ${\bf U}_l=\{({\bf X}_l,{\bf A}_l)\}$ by Eq. (\ref{eqn:13}). ${\bf X}_l\in\mathbb R^{(c\times d)}$ represents the node features.
$c'$ is the channel number of ${\bf F}$. It may not equal to the number of categories $c$.
\begin{equation}
   \setlength{\abovedisplayskip}{3pt}
   \setlength{\belowdisplayskip}{3pt}
    {\bf X}_l={{\bf \Omega}_l}_1 \times \varphi_l({\bf F}) \times {{\bf \Omega}_l}_2,
    \label{eqn:13}
\end{equation}
where ${{\bf \Omega}_l}_1\in \mathbb R ^{c\times c'}$ and ${{\bf \Omega}_l}_2\in \mathbb R ^{(w'h')\times d}$ represent the projection matrices, and $\varphi_l ({\bf F})$ represents reshaping the matrix ${\bf F}$ from $\mathbb R^{c'\times w'\times h'}$ to $\mathbb R^{c' \times (w'h')}$.
Then we perform graph reasoning similar to Eq. (\ref{eqn:2}) to obtain the graph model ${\bf Z}_l\in \mathbb R ^{c\times d}$.
Note that the way to build the graph model in LCM is different from the way to GSM in \ref{section_3.2}.

We convert the graph feature ${\bf Z}_l$ to the weight of the feature maps, and perform the element-wise product of the weight and the input feature map ${\bf F}$ to get the spatially and locally rectified feature map ${\bf F}_{rectified}$ by Eq. (\ref{eqn:14}),
\vspace{-0.1cm}
\begin{equation}
   \setlength{\abovedisplayskip}{3pt}
   \setlength{\belowdisplayskip}{3pt}
   \renewcommand\arraystretch{1.4}
\left\{
   \begin{array}{l}
   {\bm{\theta}}_l= softmax(GAP({\bf Z}_l)) \\
   {\bf F}_{rectified} = {\bm \theta}_l \odot {\bf F}
   \end{array}
\right.
   \label{eqn:14}
\end{equation}

By local graph reasoning, we capture the contexts and get the spatial relation of pixels indirectly.
Because of the structural nature of human parts, we utilize the global information ${\bf Z}_g$ to adjust the local information ${\bf Z}_l$ for better local category consistency, and the first equation in Eq. (\ref{eqn:14}) is transferred to
\begin{equation}
   \setlength{\abovedisplayskip}{3pt}
   \setlength{\belowdisplayskip}{3pt}
    {\bm{\theta}}_l= softmax(GAP({\bf Z}_l +\alpha {\bf Z}_g))
    \label{eqn:15}
\end{equation}
where $\alpha$ is the weight of global auxiliary.

\subsection{Graph Module Integration}
\label{section_3.5}
We correct the global structural error and the local consistency error of the predicted masks by our GSM and LCM, respectively.
The graph features of the two modules have different level of feature representation.
Thus, we can not simply integrate them by addition.
For the two types of errors, the local error is lower-level than the global one and should be corrected before the global structure error.
It is worth noting that the process is not reversible.
Thus the GSM and LCM are integrated by a cascaded way, as shown in Eq. (\ref{eqn:16}).
\vspace{-0.05cm}
\begin{equation}
\label{eqn:16}
   \setlength{\abovedisplayskip}{3pt}
   \setlength{\belowdisplayskip}{3pt}
    {\bf F}_{rectified} = {\bm{\theta}}_g\odot \phi({\bm{\theta}}_l \odot {\bf F}),
\end{equation}
where $\phi(.)$ represents multi-layer convolutions.
Then we integrate these two rectification modules to obtain stronger representation ability.

\subsection{Training Strategy}
\label{section_3.6}

Pseudo-labels for retraining may exploit abundant features of the object, making the segmentation network have stronger inference power.
However, lots of noisy pseudo-labels are introduced and the network is not capable to rectify itself autonomously, causing error accumulation and network degradation.
Simply using the originally predicted masks to retrain the segmentation network cannot enhance the performance effectively.

On the contrary, our rectification network is proposed to learn the error distribution of noisy predicted masks and reduce the global and local errors.
The training process is elaborately shown as Alg. \ref{alg:1}.

\begin{algorithm}
   \renewcommand{\algorithmicrequire}{\textbf{Input:}}
   \renewcommand{\algorithmicensure}{\textbf{Output:}}
   \caption{Training process}
   \label{alg:1}
   \begin{algorithmic}[1]
      \REQUIRE data with groundtruth: $\{ {\bf x}_l, {\bf y}_l\} $;
      data without groundtruth: $\{ {\bf x}_u\} $;
      segmentation network: $S-Net$;
      rectification network: $R-Net$.
      \ENSURE augmented data: $\{ {\bf x}_u, {\bf y}_u^{rectified}\} $;
      retrained segmentation network: $S''-Net $.
      \STATE Use labeled data $\{ {\bf x}_l, {\bf y}_l\} $ to train the segmentation network $S-Net$, and obtain $S'-Net$;
      \STATE Use trained segmentation network $S'-Net$ to predict the labeled and unlabeled data $\{ {\bf x}_l, {\bf x}_u\} $, and get pseudo-labels $\{ {\bf y}_l^{fake}, {\bf y}_u^{fake}\} $;
      \STATE Use $\{{\bf x}_l, {\bf y}_l^{fake}, {\bf y}_l\} $ to train the $R-Net$, and obtain $R'-Net$;
      \STATE Use $R'-Net$ and $\{{\bf x}_u, {\bf y}_u^{fake}\}$ to obtain ${\bf y}_u^{rectified}$;
      \STATE Use $\{ {\bf x}_u, {\bf y}_u^{rectified}\} $ and $\{ {\bf x}_l, {\bf y}_l\} $ to retrain $S-Net $, and obtain $S''-Net$;
      \STATE \textbf{return} $\{ {\bf x}_u, {\bf y}_u^{rectified}\} $, $S''-Net$.
   \end{algorithmic}
\end{algorithm}


\section{Experiments}

\subsection{Datasets and Metrics}
\paragraph{Datasets.}
We evaluated our algorithm on LIP and ATR datasets.
The LIP dataset~\cite{gong2017look} is currently the largest dataset of human parsing, containing many difficult samples, \emph{e.g.}, severely missing human parts, back-towards-lens, and complicated human posture.
It provides 50462 images, including 30462 images for training, 10000 images for verification and 10000 images for testing.
There are 20 categories labeled in the LIP dataset, involving 12 types of clothing, 7 types of human parts, and the background as one category.
The ATR dataset~\cite{liang2015deep} contains 18 categories of pixel-wise annotations, including 6 categories of body parts, 11 categories of clothing and the background.
It has 17700 images, including 16000 for training, 1000 for testing and 700 for verification.
\vspace{-0.3cm}

\paragraph{Metrics}
We follow the rules in the protocol of the LIP dataset to use pixel-accuracy, mean-accuracy and mean IoU as the evaluation criteria.
We use the criteria of pixel accuracy, foreground accuracy, average precision, average recall and average F1-score for the ATR dataset.

\subsection{Implementation Details}
The baseline of our segmentation network is the CE2P algorithm~\cite{ruan2019devil}, and the backbone of our rectification network is ResNet-101~\cite{he2016deep}, since our graph reasoning is performed at a high semantic level.
The size of input images is $384\times384$.
We adopt data augmentation in training and retraining, like random scales (0.5 to 1.5), cropping and horizontal flipping.
We both train and retrain our networks for 150 epochs in experiments under the semi-supervised setting.
In the optimization process, we adopt SGD optimizer with momentum of 0.9 and weight decay of 5e-4.
We use the ``poly'' learning rate strategy with an initial rate of 0.007.
The node number of low-level graph model {$c$} is a hyper-parameter, and $c$=\{20, 18\} for \{LIP, ATR\} dataset.
We set the batch size as 10 per-GPU.
\begin{table}[tb]
\centering
\footnotesize
\scalebox{0.9}{
\begin{tabular}{c|ccc|ccc|c|c}
    \toprule
    DS      &R          &G          &L             & P-Accu& M-Accu & M-IoU & IoU.I. & IoU.D.\\
    \midrule
    \midrule
      &           &           &                 & 82.10& 51.00& 39.79& 0.00 & 13.31\\
      &\checkmark    &           &                 & 82.66& 51.30& 40.64& 0.85& 12.46\\
    1/8  &\checkmark    &\checkmark    &                 & 83.11& 53.57& 42.32& 2.53& 10.78\\
      &\checkmark    &           &\checkmark          & 83.06& 53.28& 42.18& 2.39& 10.92\\
      &\checkmark    &\checkmark    &\checkmark          & \textbf{83.26}& \textbf{54.12}& \textbf{42.79}& \textbf{3.00}& \textbf{10.31}\\
    \midrule
      &           &           &                 & 83.47& 54.75& 43.06& 0.00& 10.04\\
      &\checkmark    &           &                 & 84.21& 55.89& 44.66& 1.60 & 8.44\\
   1/4   &\checkmark    &\checkmark    &                 & \textbf{85.77}& 59.91& 48.34& 5.28 & 4.76\\
      &\checkmark    &           &\checkmark          & 85.73& 59.70& 48.24& 5.18& 4.86\\
      &\checkmark    &\checkmark    &\checkmark          & 85.72& \textbf{61.11}& \textbf{48.60} & \textbf{5.54} & \textbf{4.50} \\
   \midrule
      &           &           &                 & 85.01& 59.24& 47.00 & 0.00& 6.10\\
      &\checkmark    &           &                 & 85.57& 59.49& 47.84 & 0.84& 5.26\\
   1/2   &\checkmark    &\checkmark    &                 & 86.60& 62.47& 50.56 & 3.56& 2.54\\
      &\checkmark    &           &\checkmark          & 86.28& 62.09& 49.85 &2.85& 3.25\\
      &\checkmark    &\checkmark    &\checkmark          & \textbf{86.67}& \textbf{64.89}& \textbf{50.99}& \textbf{3.99}&  \textbf{2.11}\\
   \midrule
   1     &           &           &                 & 87.37& 63.20& 53.10 &0.00 & 0.00\\
   \bottomrule
\end{tabular}}
\vspace{0.2cm}
\caption{Semi-supervised experiments in LIP dataset. DS denotes the labeled data size.
R, G, and L respectively represent the retraining strategy, global structure module, and local consistency module.
IoU.I. denotes mean IoU increased, and IoU.D. denotes the difference between the value here and the value of using whole labeled dataset.
}
\label{table_1}
\vspace{-0.5cm}
\end{table}


\subsection{Ablation Study}
We conduct elaborate ablation experiments for our rectification strategy and the graph reasoning modules.
In section \ref{4.3.1}, we conduct experiments under semi-supervised settings to evaluate our retraining and rectification strategy.
We adopt different amount of labeled and unlabeled samples for semi-supervised learning.
In section \ref{4.3.2}, we perform more experiments under supervised setting to test our GSM and LCM.
\vspace{-0.2cm}
\subsubsection{Semi-Supervised Training Test}
\label{4.3.1}
We first conduct the semi-supervised learning experiments with different training settings.
With the rectification module consisting of the GSM and the LCM, we expand the dataset with rectified masks as pseudo-labels.
In the dataset, we consider a small number of samples with ground truth as labeled data, and the other samples without ground truth as unlabeled data.
By retraining the segmentation network with labeled samples and samples with rectified pseudo-labels, the segmentation network gets better performance.
\begin{table*}
\centering
\renewcommand\arraystretch{1.5} 
\huge
\resizebox{\textwidth}{!}{
\begin{tabular}{r|cc|cccccccccccccccccccc|c|c|c}
    \toprule
    Method & L & G & \rotatebox{90}{bkg} & \rotatebox{90}{hat} & \rotatebox{90}{hair} & \rotatebox{90}{glove} & \rotatebox{90}{glasses} & \rotatebox{90}{u-cloth} & \rotatebox{90}{dress} & \rotatebox{90}{coat} & \rotatebox{90}{socks} & \rotatebox{90}{pants} & \rotatebox{90}{j-suits} & \rotatebox{90}{scarf} & \rotatebox{90}{skirt} & \rotatebox{90}{face} & \rotatebox{90}{l-arm} & \rotatebox{90}{r-arm} & \rotatebox{90}{l-leg} & \rotatebox{90}{r-leg} & \rotatebox{90}{l-shoe} & \rotatebox{90}{r-shoe} & \rotatebox{90}{P-Accu} & \rotatebox{90}{M-Accu} & \rotatebox{90}{mIoU}\\
    \midrule
    \midrule

    \multirow{4}{*}{PSPNet\cite{zhao2017pyramid}} && & 84.88& 59.86& 66.50& 32.40& 14.40& 65.79& 33.73& 52.82& 39.04& 70.04& 27.40& 14.53& 26.93& 69.14& 54.09& 57.01& 39.78& 40.41& 27.84& 27.82& 84.77& 55.26& 45.22\\

    & \checkmark& & 85.40& 62.25& 67.07& \textbf{36.12}& 21.88& 66.42 & \textbf{34.18}& 53.45& 42.04& 71.11& 28.46& \textbf{18.35}& \textbf{29.10}& 70.02& 55.75& 58.60& \textbf{46.70}& 45.60& \textbf{32.25}& 32.31& 85.23& 59.43& 47.85\\

    & &\checkmark & \textbf{85.76}& \textbf{62.93}& \textbf{67.75}& 35.90& 21.57& 66.73& 33.88& 54.53& 40.56& 71.43& \textbf{30.70}& 15.89& 24.98& \textbf{70.46}& \textbf{56.63}& \textbf{59.72}& 44.75& 45.40& 32.02& \textbf{33.05}& \textbf{85.46}& 58.98& 47.73\\

    & \checkmark& \checkmark& 85.56&62.54& 67.04& 35.94& \textbf{23.29}& \textbf{66.91}& 33.53& \textbf{54.68}& \textbf{42.20}& \textbf{71.57}& 29.96& 17.66& 28.62& 70.25& 55.87& 58.70& 45.81& \textbf{45.86}& 32.03& 32.88& 85.42& \textbf{59.56}& \textbf{48.04}\\
    \midrule
    \midrule

    \multirow{4}{*}{CE2P \cite{ruan2019devil}}& & & 86.23& 66.07& 70.41& 37.75& 31.87& 67.16& 29.91& 55.01& 43.99& 70.66& 32.30& 19.60& 25.71& 72.46& 59.17& 62.68& 50.77& 50.12& 35.40& 35.53& 85.90& 61.57& 50.14 \\

     &\checkmark & & 87.75& 65.99& 71.53& 41.99& \textbf{31.46}& 69.25& 34.73& 56.55& 47.99& 74.76& 30.99& 22.88& 28.97& 74.05& 64.66& 67.18& \textbf{58.84}& 57.87& 45.07& \textbf{46.64}& 87.37& 65.51& 53.99\\

     & &\checkmark & \textbf{87.90}& \textbf{66.67}& \textbf{71.87}& 42.54& 30.15& \textbf{69.98}& 36.98& \textbf{57.25}& \textbf{49.18}& 74.88& \textbf{35.52}& 20.45& 28.08& \textbf{74.71}& \textbf{64.77}& \textbf{67.72}& 58.41& \textbf{58.19}& 44.10& 45.37& \textbf{87.53}& 65.89& 54.08\\

     & \checkmark& \checkmark& 87.70& 65.74& 71.55& \textbf{42.58}& 30.62& 69.44& \textbf{37.13}& 56.05& 47.34& \textbf{74.92}& 31.18& \textbf{23.77}& \textbf{30.44}& 74.70& 64.73& 67.27& 57.18& 57.89& \textbf{45.82}& 46.30& 87.35& \textbf{66.11}& \textbf{54.12}\\

    \bottomrule
\end{tabular}}
\vspace{0.02cm}
\caption{Ablation experiments in LIP dataset for different base algorithms. G, and L respectively represent the GSM, and LCM.}
\label{table_3}
\vspace{-0.2cm}
\end{table*}

{\bf The LIP dataset.} The experimental results on the LIP dataset are shown in Tab.~\ref{table_1}.
We separately adopt $\{1/8, 1/4, 1/2\}$ of the samples as labeled data, and compare the results between retraining and training phases.
When the retraining strategy is without the GSM and LCM, the performance of mIoU is only improved by $\{0.85\%, 1.60\%, 0.84\%\}$.
Although the quantity of retraining samples (the unlabeled samples with pseudo-labels) is large, the performance is not improved obviously.
The experimental results demonstrate that severely noisy pseudo-labels consisting of the global and local errors lead to error accumulation and model degradation.

On the contrary, the performance of our retraining network is improved by $\{3.00\%, 5.54\%, 3.99\%\}$ respectively in the mIoU criterion, with the assistance of GSM and the LCM.
When using GSM, the mIoU is improved by $\{2.53\%, 5.28\%, 3.56\%\}$, and when applying the LCM, the mIoU is improved by $\{2.39\%, 5.18\%, 2.85\%\}$. 
We attribute the phenomenon to the characteristic of the base segmentation network, having a weak ability to distinguish the human parts and leading to more global body-part errors than the local consistency errors.

Moreover, when we use only $1/4$ labeled samples of the dataset with retraining and rectification strategies, the performance achieves considerable improvement, by $5.54\%$ in mIoU.
And it even outperforms the result of using 1/2 labeled samples without retraining and rectification strategies.
This demonstrates the high-quality of our rectified mechanism.
For the gap between using $1/2$ labeled samples and full labeled samples of the dataset, adopting our semi-supervised strategy reduces the performance drop of mIoU from $6.1\%$ to $2.11\%$.

{\bf The ATR dataset.}
The experimental results on the ATR dataset~\cite{liang2015deep} are shown in Tab.~\ref{table_2}.
Our retraining strategy with the GSM and LCM achieves $4.42\%$ improvement over the baseline with $1/2$ samples of the dataset.
Using a single global or a local module in retraining can improve the mIoU by $3.67\%$ and $3.23\%$.
The gap between the algorithms of using all samples and $1/2$ samples (without retraining) is $6.04\%$, and the gap is reduced to $1.62\%$ when adopting our rectification strategy.
It demonstrates that the retraining strategy with our rectification module is reliable for the human parsing network to improve the representation and classification power.
\begin{table}[htb]
\centering
\footnotesize
\scalebox{0.75}{
\begin{tabular}{c|ccc|ccccc|c|c}
    \toprule
    DS& R & G & L & P-Accu & F-Accu & A-Prec & Recall & F1-S & F.I & F.D\\
    \midrule
    \midrule
      &           &           &           & 92.75& 72.62& 60.56& 67.48& 63.83& 0.00& 6.04\\
      & \checkmark   &           &           & 93.61& 75.51& 61.81& 67.17& 64.38& 0.55& 5.49\\
    1/2 & \checkmark    & \checkmark   &           & 94.29& 78.33& 65.27& 69.89& 67.50& 3.67& 2.37\\
      & \checkmark   &           & \checkmark   & 94.18& 77.81& 64.81& 69.48& 67.06& 3.23& 2.81\\
      & \checkmark   & \checkmark   & \checkmark   & \textbf{94.43}& \textbf{78.74}& \textbf{66.27}& \textbf{70.36}& \textbf{68.25}& \textbf{4.42}& \textbf{1.62}\\
    \midrule
    1   &            &           &           & 94.66& 79.74& 68.26& 71.56& 69.87& 0.00& 0.00\\
    \bottomrule
\end{tabular}}
\vspace{0.02cm}
\caption{Semi-supervised experiments of 1/2 samples in ATR dataset, evaluated by pixel accuracy, foreground accuracy, average precision, recall and F1-score.
F.I denotes F1-score increase, and F.D. denotes the gap from using the whole labeled dataset.}
\label{table_2}
\vspace{-0.3cm}
\end{table}

\vspace{-0.2cm}
\subsubsection{Module Effectiveness Evaluation}
\label{4.3.2}
We apply our global structure module and local consistency module into base segmentation networks rather than the rectification network, under the supervised learning settings to evaluate the effectiveness of our two graph reasoning modules.
We perform the ablation experiments on the LIP validation set~\cite{gong2017look}.
We adopt the CE2P and PSPNet as baselines and reproduce their results by running the source codes. 
As shown in Tab.~\ref{table_3}, the proposed graph reasoning modules almost improve all the human-part categories.
By integrating the global module and the local module into CE2P~\cite{ruan2019devil}, the performance is improved by $3.94\%$ and $3.85\%$, respectively.
As for the PSPNet~\cite{zhao2017pyramid}, the mIoU is improved by $2.51\%$ and $2.63\%$, respectively.
For the confusing categories with symmetrical structure (\emph{e.g.}, left and right arms) or similar appearance (\emph{e.g.}, the coat and jumpsuits), our algorithms predict more accurately.

\vspace{-0.1cm}
\begin{figure*}[htb]
    \centering
    \includegraphics[width = 0.98\textwidth]{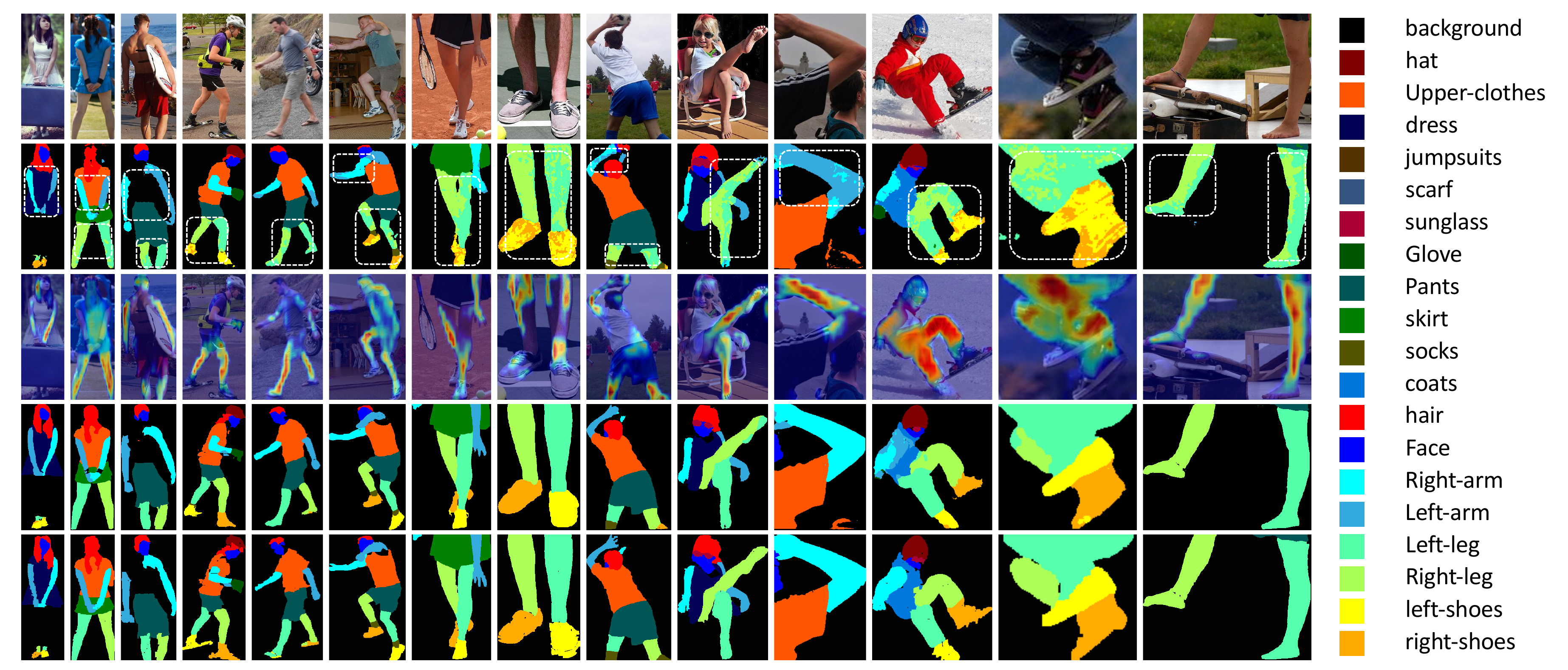}
    \caption{Visualization of the predicted masks with rectification.
    The rows correspond to the original images, the predicted masks without rectification (the errors are highlighted), the attention maps for rectification, the predicted masks with rectification, and the ground truth .}
    \label{fig:3}
    \vspace{-0.3cm}
\end{figure*}

\subsection{Compare with State-of-the-art}
We compare the proposed methods with several strong baselines on the LIP dataset, as demonstrated in Tab.~\ref{table_4}.
In the experiment setting, we treat the training set as labeled samples, and use the validation set and the test set as unlabeled samples.
According to our proposed framework, we generate pseudo-labels for unlabeled samples, then use the labeled and augmented samples to retrain the segmentation network.
As shown in Tab.~\ref{table_4}, our algorithm outperforms the other state-of-the-art algorithms. 
Specifically, our method yields a mIoU of $56.34\%$, improved by $0.44\%$ compared to the HRNet~\cite{sun2019high} and by $3.2\%$ compared to the baseline CE2P~\cite{ruan2019devil}.
Additionally, our model also outperforms the MuLA~\cite{nie2018mutual} ($49.3\%$), the  MMANs~\cite{luo2018macro} ($46.81\%$), the Attention~\cite{chen2016attention} ($42.9\%$), and the Deeplabv2~\cite{chen2017deeplab} ($41.6\%$), respectively.
MMANs~\cite{luo2018macro} adopted adversarial network with unstable and intricate training process, and MuLA~\cite{nie2018mutual} proposed to jointly train the human parsing and pose estimation networks with tremendous cost of computation and time.
But they only get limited performance in terms of mIoU.
This confirms the effectiveness of our rectification strategy.
Using unlabeled samples still makes sense for improving the representation and generalization power.

\begin{table}[tb]
\centering
\footnotesize
\scalebox{0.9}{
\begin{tabular}{r|c|ccc}
    \toprule
    Method  & Published& P-Accu & M-Accu & mIoU\\
    \midrule
    \midrule
    SegNet\cite{badrinarayanan2017segnet}   & 2017 T-PAMI       & 69.0& 24.0& 18.2\\
    FCN-8s\cite{long2015fully}              & 2015 CVPR     &76.1 &36.8 &28.3\\
    DeepLab V2\cite{chen2017deeplab}        & 2017 T-PAMI     &82.7&51.6&41.6\\
    Attention\cite{chen2016attention}      & 2016 CVPR      &83.4&54.4&42.9\\
    Attention+SSL\cite{gong2017look}       & 2017 CVPR      &84.4&54.9&44.7\\
    MMANs\cite{luo2018macro}           & 2018 ECCV       & -&-&46.81\\
    MuLA\cite{nie2018mutual}            & 2018 ECCV      &88.5&60.5&49.3\\
    JPPNet\cite{liang2018look}         & 2018 T-PAMI          & - & - & 51.37\\
    CE2P\cite{ruan2019devil}           & 2019 AAAI       &-&-&53.10\\
    HRNet\cite{sun2019high}       & 2019 ICCV           &88.21& 67.43& 55.90\\
    \midrule
    Ours & & 88.33& 66.53& \textbf{56.34}\\
    \bottomrule
\end{tabular}
}
\caption{Module effectiveness experiments in LIP dataset}
\label{table_4}
\vspace{-0.4cm}
\end{table}

\subsection{Qualitative Results}
We add our rectification network on the top of the segmentation network,
and retrain the latter one using the labeled data and the generated pseudo-labeled samples.
The visual explanation of the rectification strategy is shown in Fig. \ref{fig:3}.
In the second row of the figure, the wrongly predicted areas are highlighted in the dotted rectangles.
The CE2P algorithm correctly predicts the edges but fails to label the hard samples of semantic errors.
The third row illustrates the attention maps for rectification.
These maps are generated by calculating the gradient of the rectification network and corresponding to the original image.
It is obvious that the rectified network focuses on the mistakenly predicted human parts.
For example, for the hard samples of the back view (columns $2, 3, 4$), the samples without head (columns $7, 8, 11, 13, 14$), and the samples with complicated gestures (columns $10, 12$), the rectification network can capture the errors and perform rectification as shown in the fourth row.
These predicted errors in baseline are corrected effectively, especially for the confusing left and right limbs.
Moreover, due to the high cost of accurate pixel-wise annotation, the ground truth of the LIP dataset is low-quality in detail.
Our rectification mask performs even better than the ground truth in these details (column 8).
\vspace{-0.1cm}

\section{Conclusion}
For the problem of insufficient training data in human parsing, we managed to generate confident pseudo-labels for unlabeled samples.
We propose a rectification network for detecting and correcting predicted errors of pseudo-labels.
The rectification network consists of the global structure module and the local consistency module.
The global module is based on the hierarchical graph reasoning and captures the structural relationships of human-body parts.
The local module is based on relational local graph reasoning and captures larger semantic contexts with acceptable computational costs.
Through the rectification network, the global structural error and local consistency error of the pseudo-labels are corrected.
We retrain the segmentation network with both the labeled and pseudo-labeled samples.
Experimental results have demonstrated the effectiveness of our parsing framework.

{\small
\bibliographystyle{ieee}

}

\end{document}